\documentclass[letterpaper]{article} 
\usepackage{aaai25}  
\nocopyright
\usepackage{times}  
\usepackage{helvet}  
\usepackage{courier}  
\usepackage[hyphens]{url}  
\usepackage{graphicx} 
\usepackage{natbib}  
\usepackage{caption} 
\usepackage{algorithm}

\usepackage{newfloat}
\usepackage{listings}
\DeclareCaptionStyle{ruled}{labelfont=normalfont,labelsep=colon,strut=off} 
\floatstyle{ruled}
\newfloat{listing}{tb}{lst}{}
\floatname{listing}{Listing}
\usepackage{algorithm}
\usepackage[noend]{algpseudocode}
\usepackage{amsmath}
\usepackage{xspace}
\usepackage{xcolor}
\usepackage{amssymb}

\usepackage{subcaption}
\usepackage{amsthm}
\usepackage{multirow}
\usepackage{enumitem}
\newtheorem{thm}{Theorem}
\newtheorem{definition}[thm]{Definition}
\newtheorem{assumption}[thm]{Assumption}

\newcommand{\toolname}{AIProbe\xspace}

\title{Uncovering Systemic and Environment Errors in \\ Autonomous Systems Using Differential Testing}

\author{
    Yashwanthi Anand\equalcontrib,
    Rahil P. Mehta\equalcontrib, 
    Manish Motwani, 
    Sandhya Saisubramanian
    } 

\affiliations{
    Oregon State University
}

\begin{document}

\maketitle{}

\begin{abstract}

When an autonomous agent behaves undesirably, including failure to complete a task, it can be difficult to determine whether the behavior is due to a \emph{systemic agent error}, such as flaws in the model or policy, or an \emph{environment error}, where a task is inherently infeasible under a given environment configuration, even for an ideal agent. As agents and their environments grow more complex, identifying the error source becomes increasingly difficult but critical for reliable deployment. We introduce \emph{\toolname}, a novel black-box testing technique that applies differential testing to attribute undesirable agent behaviors either to agent deficiencies, such as modeling or training flaws, or due to environmental infeasibility. 
\toolname first generates diverse environmental configurations and tasks for testing the agent, by modifying configurable parameters using Latin Hypercube sampling. It then solves each generated task using a search-based planner, independent of the agent. By comparing the agent's performance to the planner's solution, \toolname identifies whether failures are due to errors in the agent's model or policy, or due to unsolvable task conditions.
Our evaluation across multiple domains shows that \toolname significantly outperforms state-of-the-art techniques in detecting both total and unique errors, thereby contributing to a reliable deployment of autonomous agents.
\end{abstract}

\section{Introduction}
\label{sec:introduction} 
Autonomous agents are increasingly deployed in complex real-world applications such as autonomous driving~\cite{yurtsever2020survey}, crop fertilization~\cite{gautron2022reinforcement,solow2025wofostgym}, and elderly care~\cite{bardaro2022robots,mhlanga2024artificial}. Agents operating in complex settings may sometimes produce undesirable behaviors, including failure to complete the task. We refer to such behaviors as \emph{execution anomalies}. Diagnosing the root cause of execution anomalies is critical for ensuring reliable, safe deployment.

Execution anomalies generally arise from two broad sources: (1) \emph{agent errors}: systemic errors in agent modeling or training, which results in an incorrect policy, or (2) \emph{environment errors}: unfavorable environment configuration that makes task success inherently infeasible, even for an ideal agent. Agent errors may arise from \emph{model defects} in the form of inaccuracies in the state representation, reward function, or both, in hand-crafted or learned model used for decision-making~\cite{amodei2016concrete,hadfield2017inverse,ijcai2020p50}; or \emph{training flaws} in model-free settings, such as suboptimal choices of learning algorithms or sim-to-real gaps~\cite{ramakrishnan2020blind}. On the other hand, some environment configurations are intrinsically unfavorable to agent success, such as poorly placed air vents in warehouses that reduce the efficiency of robot navigation~\cite{amazonwarehouse}, and some tasks are inherently infeasible, such as painting the wall in blue color and red color at the same time.

Consider a simple example: a mobile robot in a warehouse repeatedly fails to deliver packages from the counter to a storage area. Without a principled investigation, it is unclear whether the failure is due to (1) \emph{agent error}: the robot's policy is flawed because its model did not include information about avoiding slippery tiles, or its path planning algorithm may be suboptimal; or (2) \emph{environment error}: a newly placed pallet may have blocked all feasible paths to the goal, making the task inherently infeasible even for an optimal agent.

As both environments and agents grow more complex, identifying the source of execution anomalies becomes increasingly difficult. In practice, such anomalies are often incorrectly and reflexively attributed solely to agent errors. However, if the root cause lies in the environment configuration, no amount of training or verification will resolve the issue unless the environment itself is modified. Without a principled investigation that involves checking for alternative feasible paths using a model-independent planner or simulating task variants, it is difficult to determine whether the issue lies in the agent or the environment. 
While prior works have focused on testing for model errors~\cite{he2024curiosity,nayyar2022differential,pang2022mdpfuzz} or using formal verification methods to provide guarantees on the occurrence of anomalies~\cite{corsi2021formal,shea2024formal}, they do not determine whether an anomaly is due to the agent or the environment, \emph{without} requiring detailed internal access to the agent.

\begin{figure*}[t]
    \centering
    \includegraphics[scale=0.5]{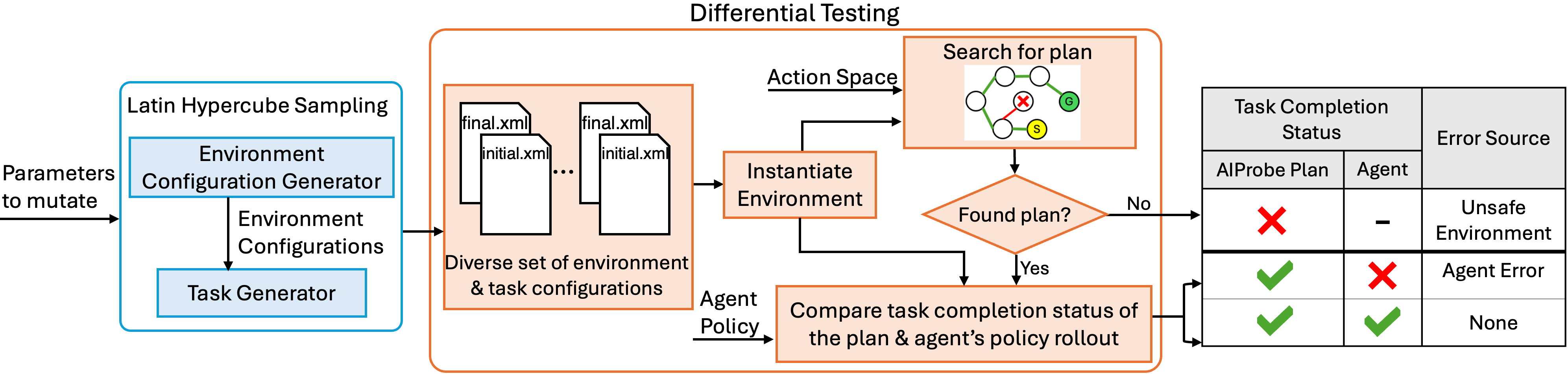}
    \caption{Overview of AIProbe}
    \label{fig:overview}
\vspace{-2ex}
\end{figure*}

We present \emph{\toolname}, a black-box technique that applies differential testing to determine whether execution anomalies are due to agent deficiencies or environment-induced infeasibility. Differential testing is a software testing methodology in which the same inputs are run through two or more independent systems (or solvers) and their outputs are compared~\cite{mckeeman1998differential}. If the outputs differ, it indicates error in one of those systems. \toolname does not require access to the agent's internal model or training data, treating the agent as a black-box system.  To uncover configurations that are unfavorable for agent deployment, \toolname systematically generates diverse environment configurations, by modifying the configurable parameters of a base environment (e.g., repositioning objects), using Latin Hypercube Sampling (LHS)~\cite{loh1996latin}. For each generated configuration, \toolname generates a range of tasks and tests whether the agent can complete them. 

To determine which tasks are feasible in an environment, \toolname uses an independent search-based oracle planner that shares the agent's action space but not its model. The planner aims to find a satisficing sequence of actions to solve these tasks. By comparing the agent's behavior with that of the oracle planner, \toolname determines whether the anomaly stems from the agent's decision-making or from the infeasibility of the task itself. If the agent fails a task that \toolname can complete safely, this suggests an agent error. If no safe plan exists, the environment itself is unsuitable for task completion. Figure~\ref{fig:overview} illustrates our approach. Note that \toolname \emph{does not localize} the modeling or training step that causes the error, nor does it reason why the task is unsolvable, and this is by design.  We take the pragmatic stance that having a principled approach to perform such a high-level diagnosis is a prerequisite for a more fine-grained error localization. 

By comparing agent behavior with planner performance across a large suite of environment-task instances, \toolname can automatically detect model flaws, environment-induced infeasibility, and edge-case behaviors, thereby identifying the operational boundary of safe deployment of autonomous systems. This analysis is also critical for reliability assessment~\cite{olamide2020autonomous} and to generate model cards for autonomous systems~\cite{mitchell2019model}. 
Our evaluation on five domains shows that our black-box differential testing method outperforms the state-of-the-art methods in error detection, in both discrete and continuous settings.

\section{Problem Formulation}
In goal-oriented sequential decision-making tasks, an agent must optimize a sequence of decisions to achieve a goal in an environment. The environment is modeled as a Markov decision process (MDP), formally defined by the tuple $M=\langle S, A, T, R,s_o,s_G\rangle$, where $S$ is a set of states, $A$ is the set of all actions that an agent can take, $T: S \times A \rightarrow S$ is the deterministic transition function determining the successor state when taking an action $a \in A$ in state $s \in S$, $R: S \times A \rightarrow \mathbb{R}$ specifies the reward associated with taking an action $a \in A$ in the state $s\in S$, $s_0 \in S$ and $s_G \in S$ denote the agent's start and goal states. 
We focus on both model-free and model-based decision-making settings. In the model-free setting, a reinforcement learning (RL) agent \emph{learns} a policy by exploring the environment. In the model-based setting, the agent \emph{computes} a policy, using its model of the environment--- either learned by exploring the environment or prescribed by an expert.

\paragraph{Problem Statement} Given a set of possible environment configurations $\mathcal{E}$, an agent with policy $\pi$ obtained either through training in a simulator or by solving its MDP $M$, and a baseline oracle planner that computes $\pi_b$ to solve a task $Z$ in an environment $E \in \mathcal{E}$, our goal is to distinguish between anomalies arising from infeasible tasks, where no policy can succeed under the given environment configuration, and those resulting from defects in the agent's model or training practices.

\begin{assumption}
    [Black-box agent access] We treat the agent as a \textbf{black box}: we can provide it with a task and observe its behavior, but we do not assume access to its policy, model, or learning process.
\end{assumption}

\begin{assumption}
    [Simulator access] We assume the agent's behavior and the oracle planner's output can be determined using a simulator.
\end{assumption}

The availability of such agents with simulators is a common assumption as most AI systems already use simulators for training. 

\paragraph{Execution Anomalies} An execution anomaly is any undesirable behavior such as the agent going around in cycles without reaching the goal state, entering a failure terminal state (such as a crash), or stepping into unsafe undesirable states (such as breaking a vase). Such behaviors may be due to agent errors, or unfavorable environment and task configurations that are fundamentally impossible to achieve.  We consider three sources of agent errors: (1) \textit{inaccurate state representation}: missing key features required for decision making; (2) \textit{inaccurate reward function}: does not fully capture the desired and undesired behaviors of the agent; or (3) \textit{both}: inaccurate state representation and reward function. Such defects lead to incorrect policy  both in model-based decision-making and in model-free settings since the agent learns a policy directly often by training in a simulator that is prone to these defects. We do not consider anomalies due to external influences such as adversarial attacks.

\subsection{Task and Environment Representation} 
We now describe the task and environment representation that is used by our approach for error detection. 

\paragraph{Task} We define a task $Z$ as a goal-directed specification within an environment. Each task is characterized by an initial state and a final state. A task in an environment is considered to be \textit{solvable} if there exists at least one sequence of actions that can achieve the goal under the given environment configuration. 

\begin{figure}[t]
\small
\begin{lstlisting}
<Environment id="" type="">
	<Attribute> ... </Attribute>
	...
	<Objects>
		<Object id="" type="">
		      <Attribute> ... </Attribute>
		      ...
		</Object>
		...
	</Objects>
	<Agents>
		<Agent id=""  type="">
		      <Attribute> ... </Attribute>
		      ...
		</Agent>
		...
	</Agents>
</Environment>
\end{lstlisting}
\caption{The XML template used to represent environment configurations. It includes the environment's, objects', and agents' attributes using the ``Attribute'' template shown in Figure~\ref{fig:attribute-template}.}
\label{fig:env-template}
\vspace{-3ex}
\end{figure}

\paragraph{Environment} An environment configuration is an instantiation of tunable parameters (e.g., obstacle layout, friction coefficients, visibility range) that define a particular task instance. Diverse environment configurations can be generated by tuning the attributes of the environment, the attributes of the objects that can exist in that environment, and the attributes of one or more agents that may interact with each other and the objects in that environment.    
Figure~\ref{fig:env-template} shows the formal representation of an environment in the form of an XML~template used by \toolname. This structured representation enables exploring the space of possible configurations in a \emph{principled} manner to generate diverse configurations. 

The template represents the environment attributes (Lines~1--3 in Figure~\ref{fig:env-template}), one or more types of objects and their attributes (Lines~4--10 in Figure~\ref{fig:env-template}), and one or more agents' 
and their attributes (Lines~11--17 in Figure~\ref{fig:env-template}). Attributes are specified using a generic \emph{Attribute} template shown in Figure~\ref{fig:attribute-template}. 
For each attribute, the template captures its name, natural language description, data type, and the current value (Lines~2--5 in Figure~\ref{fig:attribute-template}).

Since some attributes may stay constant (e.g., gravitational force of a planet when simulating a rover), the template provides \texttt{<Mutable>} tag (line~6 in Figure~\ref{fig:attribute-template}) that can be set to \texttt{false} or \texttt{true} depending on if the attribute's value should remain constant or not, respectively.
The \texttt{<Constraint>} tag (Line~7 in Figure~\ref{fig:attribute-template}) describes the constrains on values that the attribute can take. 
Since the attribute can be either numerical or categorical, this tag allows users to specify the range or categories of values that attribute can take. 
The \texttt{range} and \texttt{categories} can be described both in terms of constants or using formulas that reference the other attributes (e.g., an agent's coordinates ($x,y$) depend on the size of grid ($grid\_size$), which is represented as \texttt{<Constraint Range=[1,~{$grid\_size$}]>}). 
Finally, the \texttt{NumValues} describes the number of values that the attribute takes, which is useful to represent attributes of array or list data types (e.g., $ground\_types$ attribute may denote a sequence of floor heights (``0'', ``1'', ``2'') that for a stretch of land on which an agent is trained to walk).

\begin{figure}[t]
\small 
\begin{lstlisting}
<Attribute>
	<Name value=""/>
	<Description value=""/>
	<DataType value=""/>
	<CurrentValue value=""/>
	<Mutable value=""/>
	<Constraint Range="" Categories="" NumValues="" />
</Attribute>
\end{lstlisting}
\caption{The XML template used to represent attributes of environment, objects, and agents along with their interdependent constraints.}
\label{fig:attribute-template}
\vspace{-2ex}
\end{figure}

\begin{definition}
    An environment-task configuration is \emph{unsafe} or unfavorable if the task is unsolvable by any sequence of agent actions in the given environment. 
\end{definition}

\subsection{Running Example}
\begin{figure}[t]
    \centering
    \includegraphics[scale=0.45]{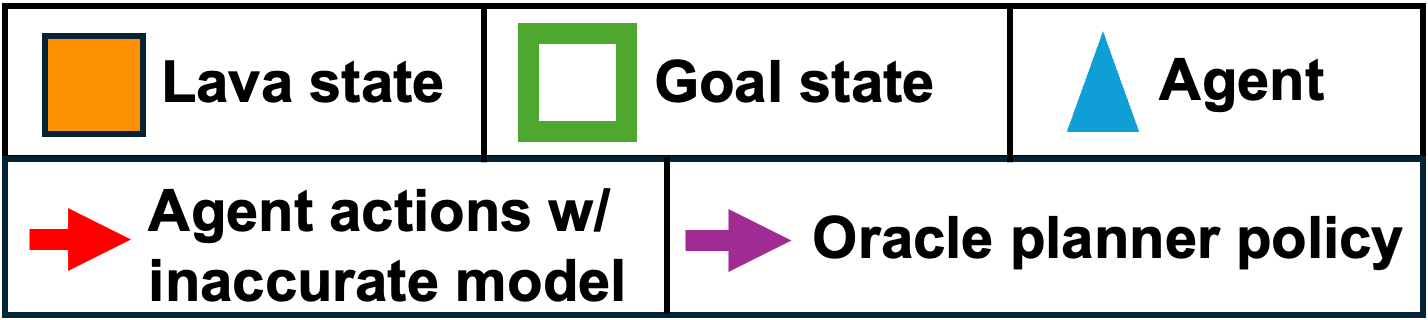} \\
    \begin{subfigure}{0.46\columnwidth}
        \centering
        \includegraphics[scale=0.45]{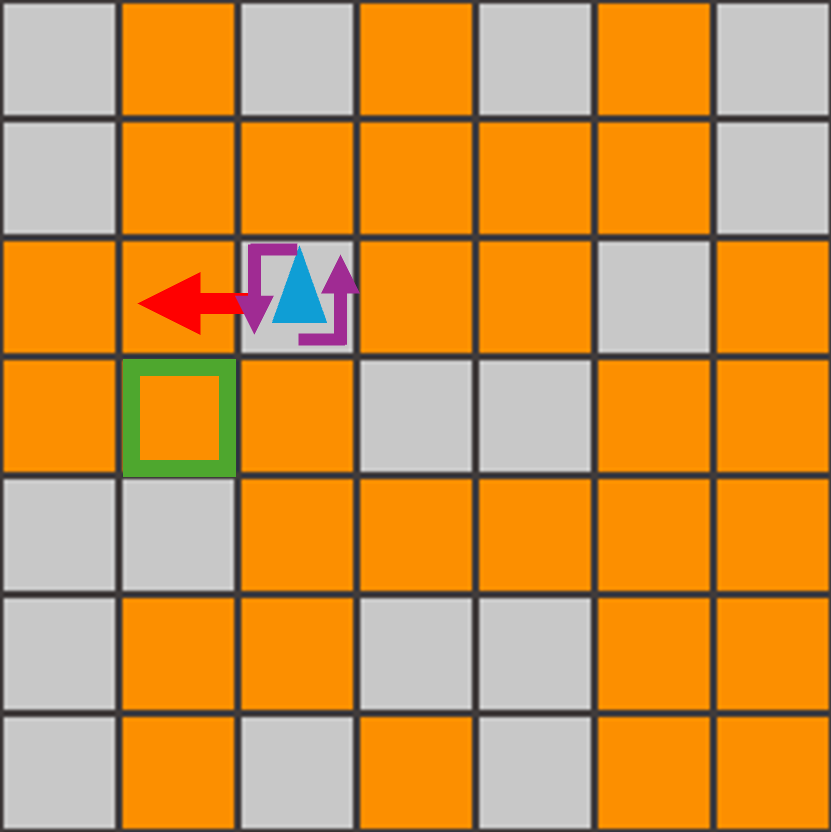}
        \caption{Unfavorable Configuration}
        \label{fig:unsafe_env}
    \end{subfigure}\hfill 
    \begin{subfigure}{0.46\columnwidth}
        \centering
        \includegraphics[scale=0.45]{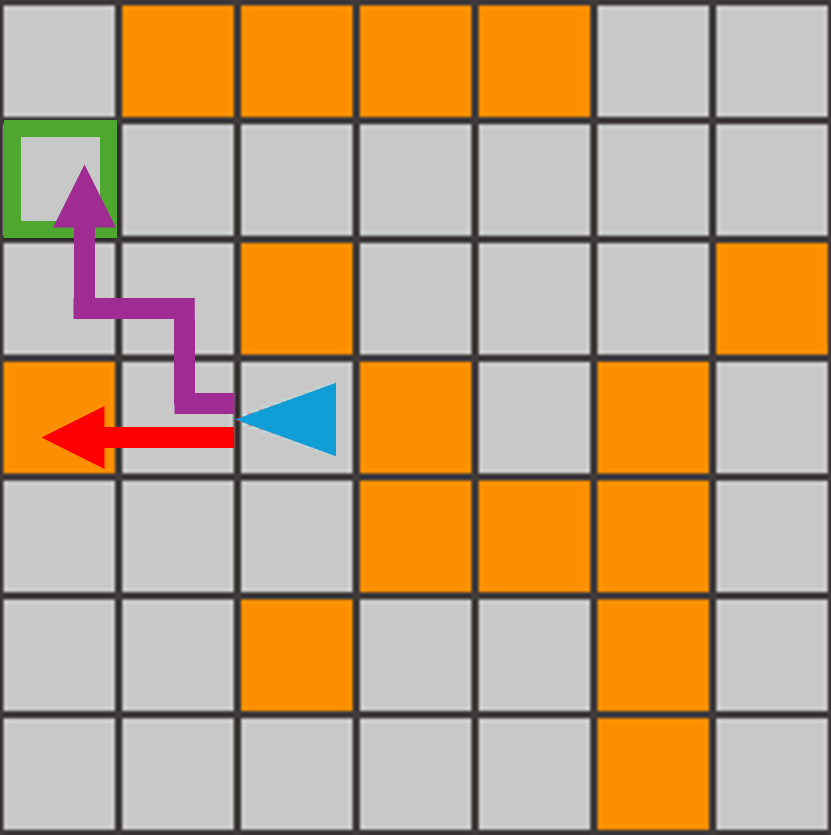}
        \caption{Favorable Configuration}
        \label{fig:inacc_r}
    \end{subfigure}
    \caption{Lava domain illustration. (a) Agent w/ an inaccurate model terminates by encountering a lava state while an agent w/ an accurate model stays in the same state stuck in a loop. (b) Agent w/ an inaccurate model fails to complete a task while the agent w/ an accurate model find a optimal path to the goal.
    }
    \label{fig:motivating_eg}
\vspace{-3ex}
\end{figure}

\begin{figure*}[t]
    \centering
    \includegraphics[scale=0.66]{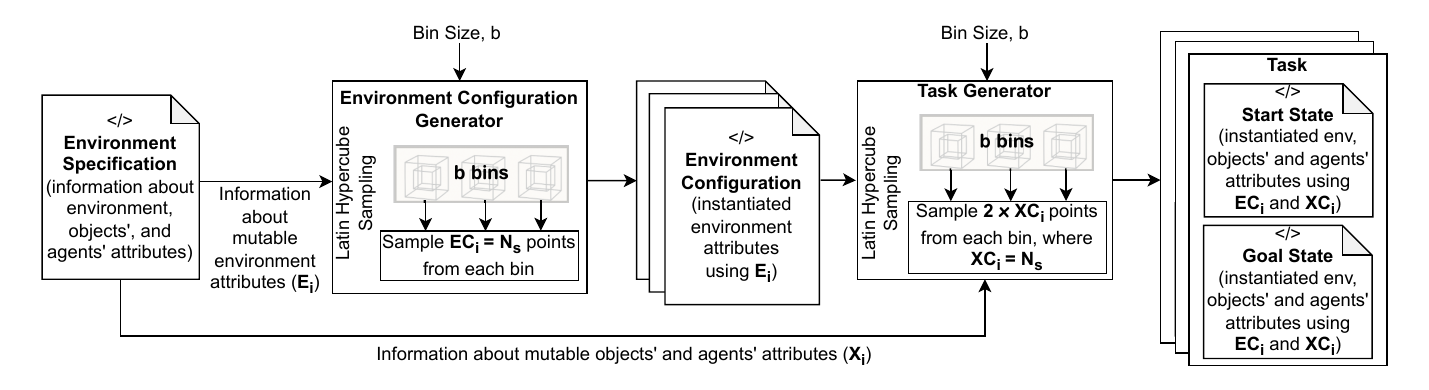}
    \caption{Generating diverse environment and task configurations uniformly at random using Latin Hypercube sampling.}
    \label{fig:lhs}
\vspace{-2ex}
\end{figure*}

Figure~\ref{fig:motivating_eg} illustrates the lava domain in which an agent must navigate to the goal location (green cell) while avoiding the lava states~\cite{MinigridMiniworld23}. This popular environment is modular and configurable, facilitating the generation of multiple configurations. Figure~\ref{fig:unsafe_env} illustrates a scenario with environment errors. The environment configuration makes it impossible to reach the goal, while avoiding the lava state. It is an example of an environment-task configuration that is unsafe or unfavorable for agent operation. Figure~\ref{fig:inacc_r} shows a setting where a path exists to the goal state but the agent steps into the lava state since it is operating based on an inaccurate model. An inaccurate model for this domain may lack information about lava in the state representation, may not penalize (enough) the agent for stepping into a lava cell, or may have a combination of both.

When an agent is unable to reach the goal, we want to automatically distinguish between the cases represented in Figures~\ref{fig:unsafe_env} and~\ref{fig:inacc_r}. As agent architectures grow more complex, especially with learned models, they end up being considered as a black-box model by the evaluators~\cite{nayyar2022differential}. Error detection in a black-box system deployed in large, complex environments is particularly challenging, as we cannot directly inspect if the agent's model accounts for lava states or manually analyze the environment-task configuration.

\section{The \toolname Approach}
Our approach operates in three phases to identify the anomaly source (Figure~\ref{fig:overview}): (1) generate diverse environment and task configurations; (2) identify feasible settings using a search-based baseline oracle planner that is independent of the agent; and (3) simulate agent behavior in feasible settings and conduct differential analysis between the observed agent behavior and expected behavior. We now describe each phase in more detail.

\subsection{Diverse Environment and Task Configurations}

Assessing variability in agent performance in different, possible configurations of the environment is critical to identify settings that are safe for agent operation. However, testing across all possible environment configurations is often practically infeasible due to the large number of possible configurations, each characterized by a large state space. To ensure broad coverage of the space of possible tasks and environments in which agents can be deployed, \toolname uses Latin Hypercube Sampling~(LHS)~\cite{loh1996latin} which is a sampling method that divides each input parameter range into equal intervals and samples within those intervals without overlap. For a D-dimensional environment, where a dimension can be continuous or discrete and may take more than one value, \toolname samples $b$ points in the space using LHS, where $b$ denotes the bin size. Thus, LHS can efficiently explore high-dimensional input spaces by ensuring that each dimension is uniformly sampled across its range. Figure~\ref{fig:lhs} shows an overview of the process. 

\paragraph{Generating Environment Configurations}
Given the manually-created XML file of the environment, such as in Figure~\ref{fig:env-template}, \toolname parses it to extract all the environment attributes, indicated by the \texttt{<Attribute>} tag. Each attribute is treated as a dimension and the mutable parameters are identified using the \texttt{<Mutable>} tag, which then forms a D-dimensional space. The input to our sampling algorithm is the number of bins~($b$), the number of dimensions~($D$), the specific details of each dimension~($EX_{i}$), such as a range for continuous dimensions or a list of categories for categorical dimensions, and the number of values per dimension~($EC_{i}$) required to represent the state of the environment. The output is a set of $b$ samples, where each sample is a point in the $D$-dimensional space. 

The algorithm iterates over each dimension $EX_{i}$. For continuous dimensions, it divides the range of $EX_{i}$ into $b$ equal bins or strata, and samples $EC_{i}$ points uniformly from each stratum. For categorical dimensions with $k$ categories, the algorithm first maps the categories to the range $[0,1]$ by partitioning the interval into $k$ equal segments. It then stratifies $[0,1]$ into $b$ equal strata, samples $EC_i$ values uniformly from each, and maps the samples back to the original categories using inverse mapping. For example, in the lava domain, environment has two attributes: grid size \emph{$n\times n$} and number of lava tiles in it denoted by \emph{$l$}. The values of $n$ and $l$ are in the ranges $[3, 50]$ and $[0, n^2]$ respectively. \toolname generates $b$~($b=100$)~diverse environment configurations of varying grid sizes and number of lava tiles by uniformly sampling across both dimensions. 
This sampling process results in $b$ environment configurations, represented by XML files with environment attributes initialized with the sampled values.

\paragraph{Generating Tasks} For each generated environment configuration, \toolname generates diverse tasks, each defined by a pair of states (start and goal) with varying object and agent attributes, while the environment attributes remain fixed. Latin Hypercube Sampling (LHS) is used to sample data points in the attributes of objects and agents that are mutable within a given environment configuration. \toolname generates two samples per bin in each mutable dimension, corresponding to initial state and final state. For example, in the lava domain, an environment configuration may specify a grid size of $5 \times 5$ with $10$ lava tiles. Each lava tile is treated as an object, and its coordinates along with the agent's start and end positions, are all considered mutable dimensions. This ensures that the placements of lava tiles and the agent's start and end positions differ across tasks. 

For each generated task, the agent is expected to plan and perform actions to move from the initial state to the final state. Since the task generation method can produce both feasible and infeasible tasks, \toolname first checks whether the task can be performed using an agent-agnostic algorithm.

\subsection{Search-based Planning as a Baseline Oracle} A natural way to determine if a task is solvable is to formulate it as a search problem to find a satisficing solution, independent of the agent's transition function or the reward function. To achieve this, \emph{any} search algorithm such as Breadth First Search (BFS) or Depth First Search (DFS) can be used as a Baseline Oracle in practice. 

Though the search is independent of the agent's model, it still requires knowledge of action effects, i.e., what happens when an action is executed in an environment. Instead of assuming privilege information about the environment dynamics, we assume the planner can pass on a sequence of actions to the simulator and observe the effect. 

\begin{assumption}
    The baseline oracle planner does not have access to privilege information: it can observe the final state reached by applying a sequence of actions from a current state, but we do not assume access to environment dynamics. 
\end{assumption}

Searching for a satsificing plan involves searching over a directed graph with states as the nodes and edges as actions. To efficiently solve problems with large search spaces, we construct the search graph on-the-fly, which complements the assumption about the lack of privilege information, and utilize a heuristic search with backtracking detailed below. 

\begin{algorithm}[t!]
\caption{Heuristic-Guided Search}
\label{alg:instruction-generation}
\small
\begin{algorithmic}[1]
\Statex \textbf{Input:} Number of bins $b$; Task $Z\!=\!(S_0, S_g)$; Action space $A$; Number of paths to explore in one iteration $N$; Maximum search depth $D$; Set of unfavorable states $S_F$

    \State $Visited \gets \emptyset$  \Comment{set of visited states}
    \State $\pi_b \gets []$ \Comment{Baseline Plan}
    \State $h \gets$ \Call{Heuristic}{$S_0$, $S_g$, $b$} 
    \State \Return \Call{Search}{$S_0, S_g, A, h, N, D, 0, Visited, \pi_b$}
\Function{Heuristic}{$S_{curr}, S_{goal}, b$}
    \State $\phi_{curr} \gets \Call{Bin\_Normalized\_Attribute}{S_{curr}, b}$
    \State $\phi_{goal} \gets \Call{Bin\_Normalized\_Attribute}{S_{goal}, b}$
    \State \Return $\|\phi_{curr} - \phi_{goal}\|_1$
\EndFunction
\Function{Search}{$S_{curr}$, $S_{goal}$, $A$, $h$, $N$, $D$, $depth$, $Visited$, $\pi_b$}
    \If{$depth > D$}                    \Comment{reached max depth}
        \State \Return (False, $\pi_b$)
    \EndIf
    \If{$S_{curr} \in S_F$}
   \If{$\pi_b = []$}
        \State \Return (False, $\pi_b$)  \Comment{terminal start state}
    \Else
        \State $(S_{prev}, \pi_b^{prev}) \gets$ \Call{Backtrack}{$S_{curr}$, $\pi_b$}     
        \State $h \gets$ \Call{Heuristic}{$S_{prev}$, $S_{goal}$, $b$}
        \State \Return \Call{Search}{$S_{prev}, S_{goal}, A, h, N, D, depth+1$, $Visited$, $\pi_b^{prev}$}
    \EndIf
    \EndIf
    \If{$(S_{curr}, \pi_b) \in Visited$}
        \State \Return (False, $\pi_b$) \Comment{Avoid revisiting states to prevent infinite loops}
    \EndIf
    \State $Visited \gets Visited \cup \{(S_{curr}, \pi_b)\}$
    \If{$S_{curr} = S_{goal}$}
        \State \Return (True, $\pi_b$) \Comment{Valid plan found}
    \EndIf
\For{$i = 1$ to $N$}
       \State $\pi_b$ = Sample $h$ actions from $A$ 
        \State $S_{next} \gets$ \Call{Transition}{$S_{curr}$, $\pi_b$} 
        \State $h \gets$ \Call{Heuristic}{$S_{next}$, $S_{goal}$, $b$}
        \State $result \gets$ \Call{Search}{$S_{next}$, $S_{goal}$, $A$, $h$, $N$, $D$, $depth+1$, $Visited$, $\pi_b$}
        \If{$result[0] = \text{True}$} \Comment{Valid plan found}
            \State \Return $result$ 
        \EndIf
    \EndFor
    \State \Return (False, $\pi_b$) 
    \EndFunction
\end{algorithmic}
\end{algorithm}

\paragraph{Depth-limited heuristic search} The input to our planner is a task~($Z = (S_0,S_G)$), with an initial state~($S_0$) and a goal state~($S_g$), agent's action space~($A$), a set of unfavorable states $S_F$ that correspond to anomalies, and the parameters used to measure the search heuristic: number of bins~($b$), number of plans to generate in one search iteration~($N$), and maximum depth along one search path~($D$). The output of the search algorithm is a \emph{satisficing} plan~($\pi_b$) to reach $S_g$, if one exists. Algorithm~\ref{alg:instruction-generation} presents the pseudocode. The key parts of our search are explained next. 

\vspace{2pt}
\noindent \emph{Heuristic Estimation:}
The heuristic value estimates how many actions are required to reach the goal from the current state, i.e., an estimate of the plan length. The heuristic estimate at each state is the L1-norm distance between the bin indices of normalized attributes of the current state and the goal state. Besides reducing the search space and guiding the search efficiently, the heuristic also aligns with our task generation using LHS, which bins each attribute to ensure task diversity. For example in the lava domain, the start and goal states differ in terms of two agent's parameters: x and y. Let a particular instance be characterized by an initial state $x=32, y=23$ and goal state $x = 13, y = 2$ in a $32\times 32$ grid. The heuristic distance between the two states, calculated using 100 bins, is $|{bin}(x=32) - {bin}(x=13)| + |{bin}(y=23) - {bin}(y=2)| = |100 - 41| + |72 - 7| = 59 + 65 = 124$. The algorithm will therefore generate a $N$ number of plans, each with $124$ actions.

\vspace{2pt}
\noindent \emph{Depth-limited recursive search:} The core of the algorithm is a recursive search procedure (Lines~9--31). The algorithm terminates when a solution has been found (Lines 22-23 and 29-30), when the maximum search depth has been reached (Line 10-11), or when the start state is a failure terminal states (such as a crash state). If an unfavorable state is reached during the search, with the current (partial) plan, then the algorithm backtracks to the previous state (roll back one step) and attempts to explore different paths~(Line 16). Revisited states are skipped. 
In each iteration, the algorithm samples $N$ action sequences, simulates them to determine the next states, re-evaluates the heuristic, and continues recursively (Lines~24-28), until a valid solution is found or all search paths are exhausted. While some generated tasks may be infeasible, the majority tend to be solvable, particularly in less constrained environments where the agent has sufficient freedom to move. For tasks where the algorithm fails to find an instruction, \toolname performs a breadth-first search from the initial state to the final state to verify that the task is indeed impossible.

\subsection{Error Attribution} Once feasible environment-task scenarios are identified, the agent's performance is evaluated on those configurations. The behavior traces of both the agent and the planner are compared using a differential analysis procedure, which in our case measures divergence in terms of task completion. If the \toolname's planner can solve the task but the agent cannot, then it is inferred that the agent's model is inaccurate. If the planner is unable to solve the task, then the environment-task setting is flagged as unsafe for agent deployment.

\section{Empirical Evaluation}
\label{sec:evaluation}
We evaluate \toolname using both discrete and continuous open-sourced, single and multi-agent domains. All reinforcement learning (RL) domains were trained using PPO~\cite{schulman_proximal_2017}. Our evaluation is driven by the following four research questions. \footnote{Code: https://github.com/ANSWER-OSU/AIProbe } 

\noindent \textit{RQ1}: How effective is \toolname in identifying execution anomalies across domains, in comparison with the current approaches?

\noindent \textit{RQ2}: How effective is \toolname in uncovering agent errors and environment errors, under different types of agent model defects?

\noindent \textit{RQ3}: (Ablation study) How much improvement can be achieved if environment-tasks configurations are generated by a large language model (LLM) conditioned on agent capabilities, instead of using Latin Hypercube sampling?

\noindent \textit{RQ4}: (Ablation study) How sensitive is \toolname to the choice of baseline planner?

\paragraph{Baselines} We compare the performance of \toolname, with 10 and 20 seeds used to generate environment-task configurations, with that of two state-of-the-art fuzz testing approaches designed specifically to test autonomous systems: MDPFuzz~\cite{pang2022mdpfuzz} and CureFuzz~\cite{he2024curiosity}. We also perform two ablation studies on components that are critical to \toolname: environment-task generation and baseline planner. Specifically, 
we compare \toolname's performance to using GPT-4o for environment-task generation, and our proposed heuristic-search baseline planner with that of Breadth First Search (BFS).

\paragraph{Evaluation Metrics}
We use the following metrics in our experiments: 
(1) the number of \emph{execution anomalies} identified, (2) \emph{environment errors}: the number of anomalies that occur due to infeasible tasks, (3) \emph{agent errors}: the number of anomalies that occur due to defects in agents (due to its model, training, or solver), and (4) \emph{state coverage} which measures the coverage of our environment-task generation. This metric is inspired by traditional software testing techniques that use code coverage ratio to demonstrate the effectiveness of the generated tests~\cite{Motwani19}. To calculate state coverage for continuous and high-dimensional state spaces, we use the same binning strategy that we apply for generating environment-task configurations. Specifically, we divide each dimension of the state space into 100 bins (bin size = 100), creating a structured grid where each bin represents a discrete state. The state coverage is then computed as the fraction of unique bin combinations generated by each technique over the total possible combinations, given by $100^D$, where $D$ is the number of dimensions in the domain. An exception is the BipedalWalker domain, where we follow the same approach as CureFuzz and compute coverage based on the proportion of unique ground types encountered over all possible ground types.

\subsection{Domains}
We use five domains for evaluation: ACAS Xu, Cooperative Navigation, Bipedal Walker, Flappy Bird, and Lava. The first three domains are commonly used by the existing approaches for evaluation; Flappy Bird represents a popular RL benchmark, and Lava provides a simple discrete environment.
In each domain, we evaluate using the \emph{base model}, which is the publicly available pre-trained model similar to existing works~\cite{he2024curiosity}, and three additional variants that we create to specific model errors: an incomplete state representation, an incorrect reward function, and a combination of both. The incomplete state representation denotes the scenario where the agent is reasoning at an abstract level that does not fully capture the details for successful task completion. Incorrect reward denotes scenarios where the under-specified reward function does not capture the full range of desirable and undesirable behaviors. 

\paragraph{ACAS Xu} This domain simulates an aircraft collision avoidance system, with two aircraft: ownship (agent) and intruder~\cite{julian2016policy}. We follow the base model design prescribed in~\citeauthor{he2024curiosity}(\citeyear{he2024curiosity}) with continuous states and discrete actions. The state representation is denoted as $\langle \rho, \theta, \psi, v_{own}, v_{int}\rangle$ where $\rho$ (m) is the distance from ownship to intruder, $\theta$ (rad) is the angle to intruder relative to ownship's heading, $\psi$ (rad) is the intruder's heading relative to ownship, $v_{own}$ (m/s) is the ownship speed, and $v_{int}$ (m/s) is the intruder speed. The agent's available actions are Clear-of-Conflict, weak left, strong left, weak right, and strong right. The reward function in the base model is given by $(\gamma + \rho / 60261.0)$ for every step, and $-100$ when the distance between the two aircrafts  is less than a certain threshold. 
We create three types of erroneous agent models: (1) an incorrect state representation that omits the distance feature $\rho$, which impairs the agent's ability to reason about the relative distance between itself and the intruder; (2) an under-specified reward function $(\gamma + \rho / 1e6.0)$ that disproportionately emphasizes the distance component, potentially leading to unsafe conditions; and (3) a combination of both incorrect state representation and reward function. 

\paragraph{Co-operative Navigation (Coop-Navi)} This is a multi-agent, continuous domain from Gymnasium's PettingZoo suite~\cite{lowe2017multi}. We consider a setting with three agents that must coordinate to occupy three distinct landmark positions while avoiding collisions with each other. Each agent can choose from five discrete actions: move left, move right, move down, move up, or take no action. The domain is characterized by continuous states and actions. In the base model, a state is represented as a list of three tuples, each corresponding to the observation of one agent. A tuple is represented as $\langle v_s, p_s, p_l, p_o, c \rangle$, where $v_s$  is the agent's velocity, $p_s$ is the agent's position, $p_l$ is the position of the three landmarks, $p_o$ is the position of the other two agents, and $c$ is a 2-bit communication channel. The agents receive a shared global reward on the sum of distances between each landmark and its nearest agent. Additionally, each agent is penalized locally with a reward of $-1$ for every collision with another agent. An episode terminates if the agents collide five times before getting close to the landmarks. Agent model errors in this domain are introduced as follows: (1) incorrect state representation that omits the positions of the other agents $p_o$ from each agent's observation, making it more difficult to coordinate; (2) an under-specified reward that does not include a penalty for collision, which may lead to potentially unsafe behaviors; and (3) both incorrect state representation and under-specified reward. 

\paragraph{Bipedal Walker} This is a continuous control RL domain where a four-joint bipedal robot learns to walk across a challenging terrain, while maximizing the number of timesteps it can stay upright without falling~\cite{brockman2016openaigym}. This domain is \emph{non-deterministic} due to its reliance on the Box2D Physics engine which is not fully deterministic and can have subtle randomness in actions. 
In our experiments, we evaluate the agent under the \emph{hardcore} setting, where the terrain can be grass, stump, stairs or a pit. An episode is successful if the agent remains upright and traverses the terrain for $2000$ timesteps, accumulating a reward of approximately $300$ points.
In the base model, the state representation includes the hull angle speed, angular velocity, horizontal speed, vertical speed, position of joints and joints angular speed, legs contact with ground, and $10$ Lidar rangefinder measurements. To encourgae forward movement, the agent gets a penalty of $-100$ when it falls, and a penalty of $-5$ for not maintaining an upright position during the course of walking across the terrain. Agent model errors in this domain are introduced as follows: (1) incorrect state representation that omits the LiDAR values, affecting the agent's ability to reason about upcoming terrain; (2) an under-specified reward function that does not penalize for not maintaining an upright posture; and (3) both incorrect state representation and reward function.

\paragraph{Flappy Bird} In this popular RL domain, the agent (bird) must learn to navigate through the gaps between pipes, aiming to maximize its survival time~\cite{tasfi2016PLE}. The domain is characterized by continuous states and discrete actions. A state in the base model is denoted by $\langle y_b, v_b, d_{p_1}, y_{p_1t}, y_{p_1b}, d_{p_2}, y_{p_2t}, y_{p_2b}\rangle$, where $y_b$ is the agent's position along the $y$ coordinate and $v_b$ is its velocity, $d_{p_1}$ is the relative distance to the next pipe, $y_{p_1t}$ and $y_{p_1b}$ denote the $y$ position of the next pipe on top and bottom respectively, $d_{p_2}$ is the distance to the pipe after the next, $y_{p_2t}$ and $y_{p_2b}$ denote the $y$ position of the top and bottom pipe after the next. The agent can either do nothing or fly up. It receives a reward of $+0.5$ for every time step that does not lead to termination, $+1$ for passing a pipe. The game terminates when the agent flies into one of the pipes, or the upper or lower boundary of the game's frame, in which case it received $-1$. Agent model errors are introduced as follows: (1) incorrect state representation that omits the vertical positions of the upcoming upper and lower pipes ($y_{p_1t}, y_{p_1b}, y_{p_2t}, y_{p_2b}$); (2) an under-specified reward function that only incentivizes survival and penalizes for early termination, without explicitly rewarding the agent for passing a pipe, due to which the agent may fail to learn to time its flaps; and (3) both incorrect state representation and reward function.

\vspace{-0.25ex}
\paragraph{Lava} We use a modified version of the Lava domain from the Minigrid environment suite~\cite{MinigridMiniworld23}. The agent's objective is to reach the goal while avoiding lava tiles, aiming to minimize the number of steps taken (Figure~\ref{fig:motivating_eg}). We modify the domain such that the agent plans using a model in this domain, allowing us to analyze \toolname in planning settings where the agent has access to a model of the environment that it uses for planning. 
In the base model, a state is represented by the tuple $\langle x,y,d,l \rangle$, where $(x, y)$ is the agent's position in the grid, $d \in {north, south, east, west}$ indicates the agent's orientation and $l$ is a binary variable indicating the presence of lava at $(x,y)$. The agent's action space consists of three discrete actions: turn left, turn right and move forward. 
The agent receives a reward of $+100$ when reaching the goal state, a penalty of $-10$ for entering a lava state, and $0$ otherwise. Stepping into a lava tile is a failure terminal state. An inaccurate state representation omits $l$ from the state representation, introducing partial observability. An under-specified reward function rewards the agent for reaching the goal but does not penalize for stepping into a lava state. 

\begin{table*}[t]
\centering
\small
\begin{tabular}{|l|l|c|c|c|}
\hline
 &  & 
\multicolumn{2}{|c|}{\textbf{Execution Anomalies Detected}} &  \\
\cline{3-4} 
\textbf{Domain} & \textbf{Method} & \textbf{Unique} & \textbf{Total} & 
\textbf{State Coverage} \\
\hline
\multirow{5}{*}{ACAS Xu} 
& MDPFuzz   & $9.0 \pm 0.9$   & $183.0 \pm 15.9$    & $2.0e^{-6} \pm 4.0e^{-8}$ \\
& CureFuzz  & $17.0 \pm 1.5$  & $\mathbf{268.0 \pm 26.8}$   & $6.0e^{-6} \pm 1.0e^{-6}$ \\
& \textbf{AIProbe (10 seeds)}  & $\mathbf{53.7\pm 5.8}$   & $53.7\pm 5.8$    & $ \mathbf{8.1e^{-3}\pm 4.8e^{-4}}$\\
& \textbf{AIProbe (20 seeds)}  & $\mathbf{54.8\pm 5.1}$   & $54.8\pm 5.1$    & $\mathbf{8.2e^{-3}  \pm  6.8e^{-4}}$ \\
\hline
\multirow{4}{*}{Coop Navi} 
& MDPFuzz   & $52.4 \pm 11.7$   & $52.4 \pm 8.8$  & $5.0e^{-19} \pm 6.0e^{-20}$ \\
& CureFuzz  & $85.3 \pm 7.3$  & $85.0 \pm 7.3$   & $1.0e^{-18} \pm 5.0e^{-19}$ \\
& \textbf{AIProbe (10 seeds)}  & $\mathbf{139.0 \pm 12.0}$   & $\mathbf{139.0 \pm 12.0 }$  & $\mathbf{9.9e^{-9} \pm 1.1e^{-10}}$ \\
& \textbf{AIProbe (20 seeds)}  & $\mathbf{138.4 \pm 10.8}$   & $\mathbf{138.6 \pm 10.9}$  & $\mathbf{9.9e^{-9} \pm 1.4e^{-10}}$ \\
\hline
\multirow{4}{*}{BipedalWalker} 
& MDPFuzz   &$126.0\pm31.8$  &$126.0\pm31.8$ & $6.5e^{-2} \pm 2.0e^{-3}$ \\
& CureFuzz  & $658.0\pm98.3$ &$658.0\pm98.3$ &  $\mathbf{4.2e^{-1} \pm2.0e^{-2}}$\\
& \textbf{AIProbe (10 seeds)}  & $\mathbf{7880.0\pm 211.4}$ & $\mathbf{7880.0\pm 211.4}$  & $3.0e^{-3} \pm 1.0e^{-4}$ \\
& \textbf{AIProbe (20 seeds)}  & $\mathbf{7890.0 \pm 166.8}$ & $\mathbf{7890.0\pm166.8}$  & $3.0e^{-3} \pm 1.0e^{-4}$ \\
\hline 
\multirow{1}{*}{Flappy Bird} 
& MDPFuzz & $3125.0\pm1334.9$ & $\mathbf{12000.0\pm6324.6}$ & $45.9\pm17.9$\\
& CureFuzz & $1376.8\pm41.1$ & $1492.0\pm41.8$& $22.3\pm0.5$\\
& \textbf{AIProbe (10 seeds) } & $\mathbf{7277.0 \pm 135.6} $   & $7992.0 \pm 95.9$  & $\mathbf{99.9 \pm 0.3}$ \\
& \textbf{AIProbe (20 seeds)}  & $\mathbf{7188.1 \pm 240.9} $   & $7960.1 \pm 111.8$  & $\mathbf{99.9 \pm 0.2}$ \\
\hline
\multirow{1}{*}{Lava} 
& MDPFuzz  &  $2160.2\pm139.7$ & $2212.0\pm150.03$  & $3.2e^{-9}\pm1.5e^{-10}$ \\
& CureFuzz  & $\mathbf{213585.4\pm106793.2}$ & $\mathbf{214310.8 \pm107155.8}$ & $9.2e^{-7} \pm 1.1e^{-7}$   \\
& \textbf{AIProbe (10 seeds)}  & $6775.5 \pm 319.7 $   & $6815.0 \pm 334.1$  & $\mathbf{8.9e^{-5} \pm 2.4e^{-5}} $ \\
& \textbf{AIProbe (20 seeds)}  & $6704.8 \pm 303.9 $   & $6726.5 \pm 317.1$  & $\mathbf{8.9e^{-5} \pm2.1e^{-5}}$ \\
\hline
\end{tabular}
\caption{Comparison of execution anomalies discovered and state coverage, across domains with different approaches. Results are averaged across different base models in each domain. Best values in each domain are indicated in \textbf{bold}.}
\label{tab:crashes}
\end{table*}

\begin{table*}[t]
    \centering
    \renewcommand{\arraystretch}{1.2}
    \resizebox{\textwidth}{!}{
    \begin{tabular}{|c|c||c|c||c|c|c|c|}
        \hline
        & & 
        \multicolumn{2}{c||}{\textbf{Env-Task Configs}} & 
        \multicolumn{4}{c|}{\textbf{\#Agent Errors}} \\
        \cline{3-4} \cline{5-8}
         \textbf{Domain} & \textbf{\#Seeds} & \textbf{\# Feasible} & \textbf{\# Infeasible (Env. error)} & \textbf{Base Model} & \textbf{Inacc. State} & \textbf{Inacc. Reward} & \textbf{Both}  \\
        \hline
        \multirow{2}{*}{ACAS Xu} 
        & 10  & $9974.8 \pm 2.3$ & $25.2 \pm 2.3$ & $262.7 \pm 33.1$ & $119.9 \pm 18.8$ & $124.8 \pm 17.2$ & $95.9 \pm 11.4$  \\
        & 20  & $9975.1 \pm 4.3$ & $24.9 \pm 4.3$ & $268.5 \pm 34.1$ & $122.5 \pm 25.1$ & $117.5 \pm 25.3$ & $90.6 \pm 15.0$  \\
        \hline
        \multirow{2}{*}{Coop Navi} 
        & 10  & $9938.7 \pm 1.3$ & $47.8 \pm 13.2$ & $98.2 \pm 11.2$ & $4687.6 \pm 279.5$ & $3372.9 \pm 238.8$ & $8997.5 \pm 655.2$  \\
        & 20  & $9939.3 \pm 1.7$ & $46.4 \pm 14.5$ & $98.0 \pm 12.0$ & $4569.2 \pm 339.5$ & $3846.8 \pm 363.9$ & $9157.2 \pm 634.2$  \\
        \hline
        \multirow{2}{*}{Flappy Bird} 
        & 10  & $1375.4 \pm 148.4$ & $932.8 \pm 142.9$ & $268.3 \pm 71.2$ & $832.9 \pm 317.8$ & $776.0 \pm 209.4$ & $675.1 \pm 199.5$  \\
        & 20  & $1406.5 \pm 160.2$ & $885.5 \pm 157.7$ & $262.1 \pm 68.6$ & $841.9 \pm 232.4$ & $703.3 \pm 189.5$ & $645.8 \pm 189.6$  \\
        \hline
        \multirow{2}{*}{Lava} 
        & 10  & $3185.0 \pm 334.1$ & $6815.0 \pm 334.1$ & $0.0 \pm 0.0$ & $2137.8 \pm 254.4$ & $2137.8 \pm 254.4$ & $2137.8 \pm 254.4$  \\
        & 20  & $3273.5 \pm 317.1$ & $6726.5 \pm 317.1$ & $0.0 \pm 0.0$ & $2233.5 \pm 244.9$ & $2233.5 \pm 244.9$ & $2233.5 \pm 244.9$  \\
        \hline
    \end{tabular}
    }
    \caption{Average number of agent and environment errors identified by \toolname, across domains and models. ``Both'' refers to a model with known defects in state representation and reward function.}
    \label{tab:agent_env_errors}
\vspace{-2ex}
\end{table*}

\begin{figure*}
    \centering
    \includegraphics[scale=0.7]{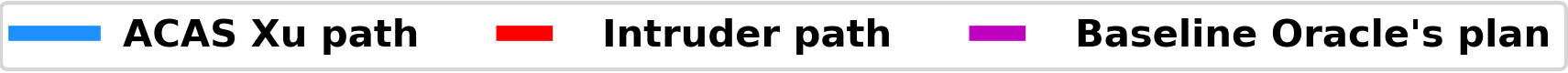}\\
    \begin{subfigure}[t]{0.24\textwidth}
        \centering
        \includegraphics[scale=0.23]{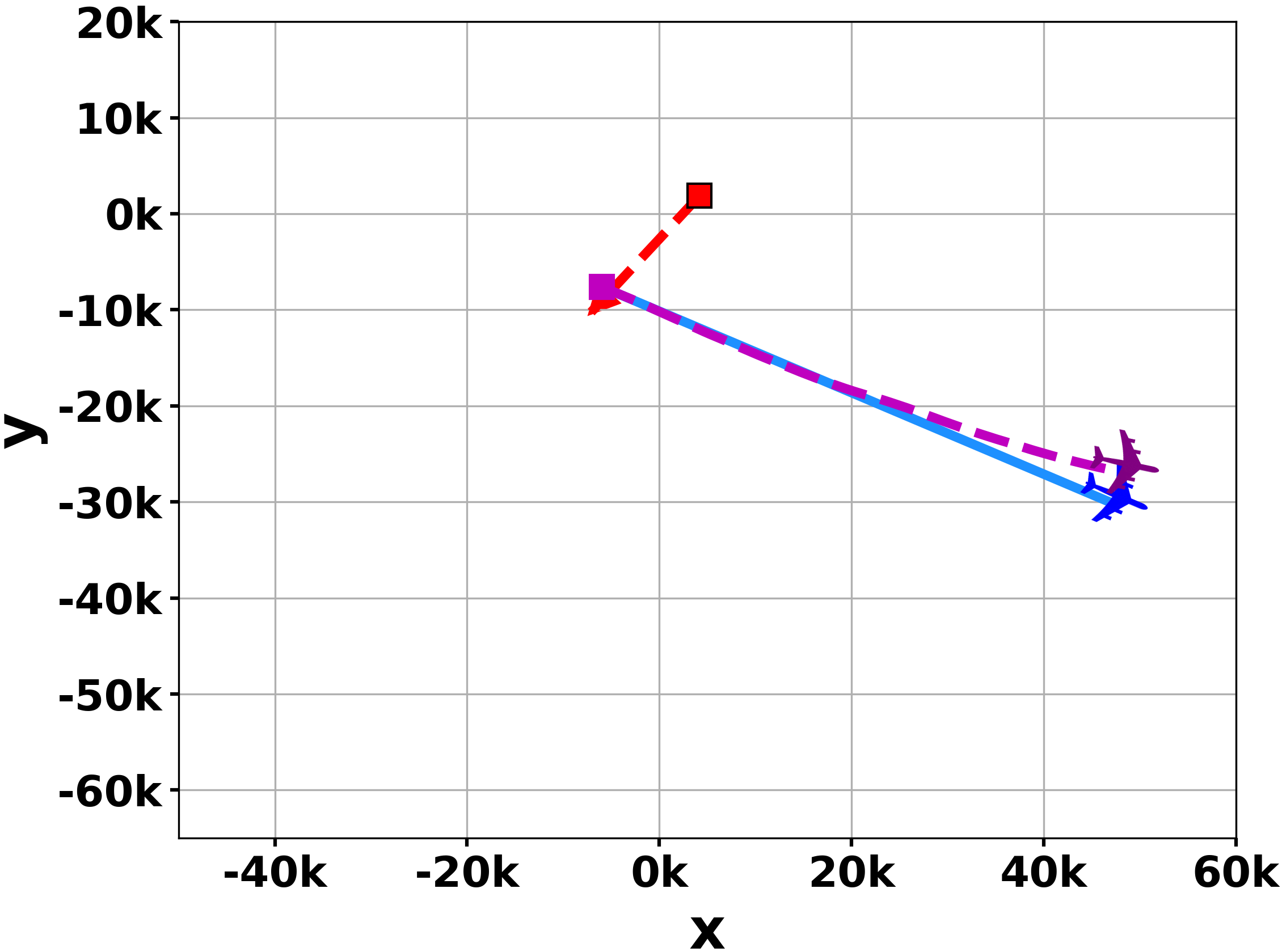}
        \caption{Base Model}
        \label{fig:baseModel}
    \end{subfigure}
    \hfill
    \begin{subfigure}[t]{0.24\textwidth}
        \centering
        \includegraphics[scale=0.23]{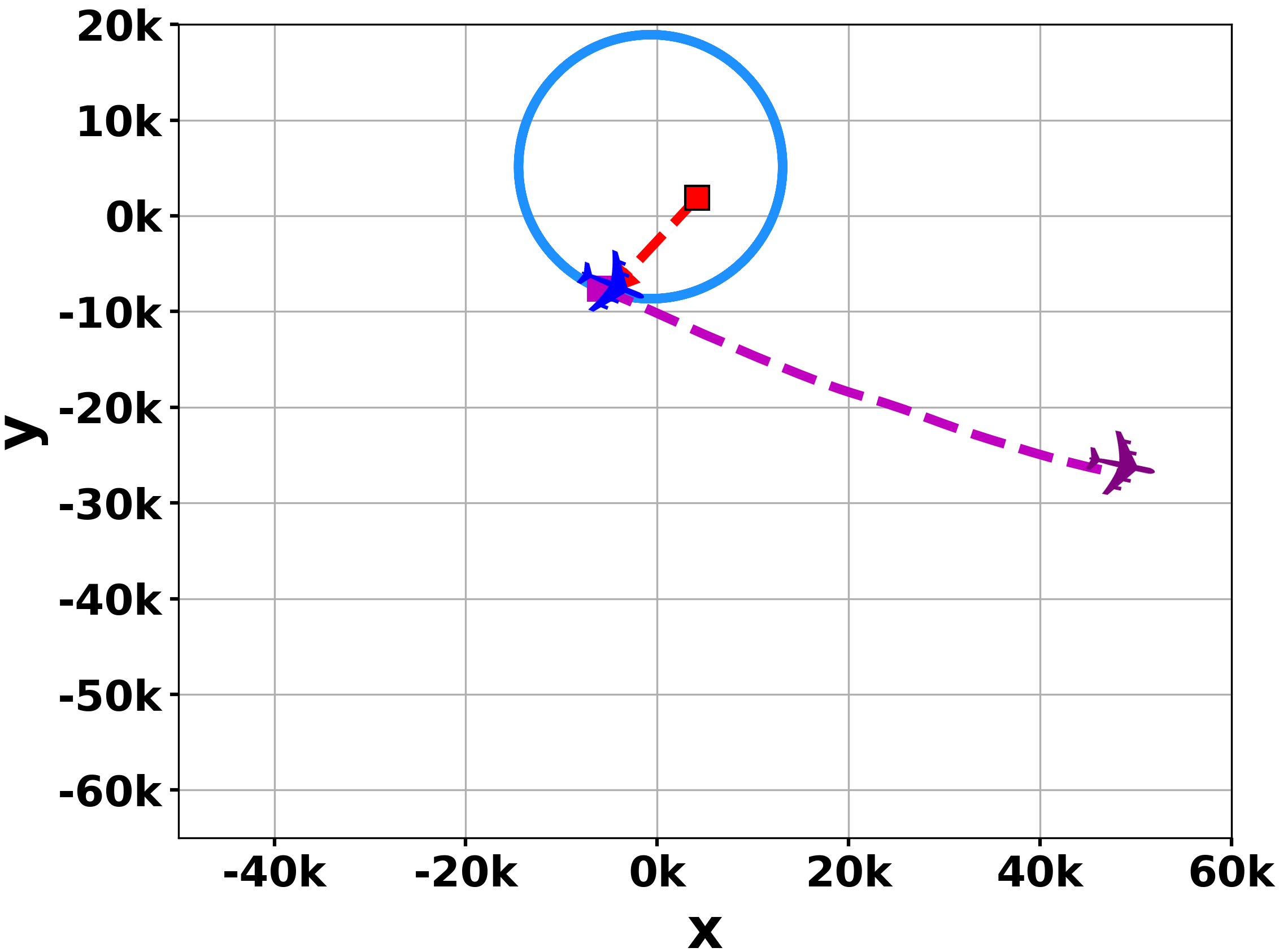}
        \caption{Model w/ inaccurate reward function}
        \label{fig:inaccReward}
    \end{subfigure}
    \hfill
    \begin{subfigure}[t]{0.24\textwidth}
        \centering
        \includegraphics[scale=0.23]{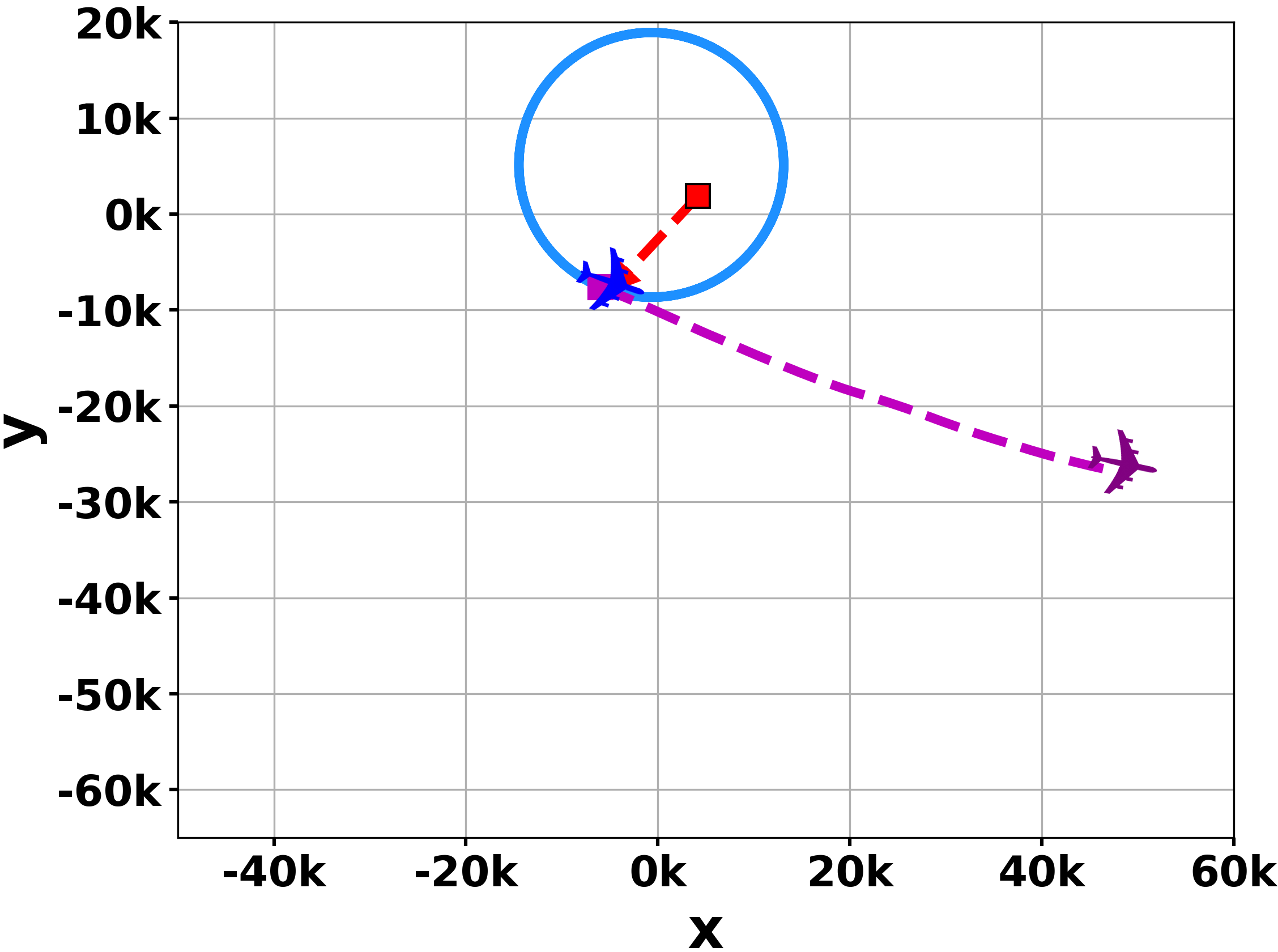}
        \caption{Model w/ inaccurate state representation}
        \label{fig:inaccState}
    \end{subfigure}
    \hfill
    \begin{subfigure}[t]{0.24\textwidth}
        \centering
        \includegraphics[scale=0.23]{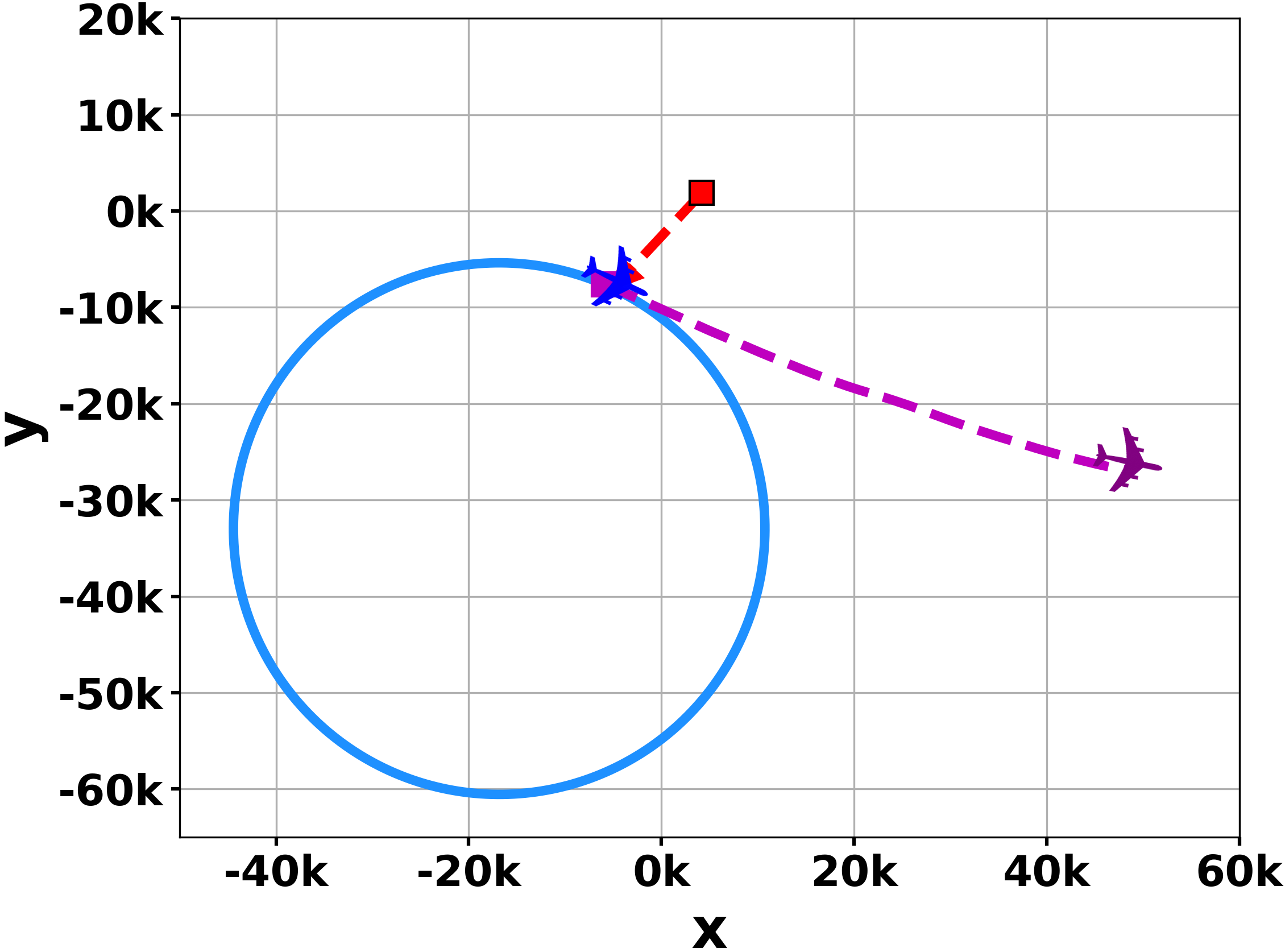}
        \caption{Model w/ inaccurate reward function \& state representation}
        \label{fig:inaccRewardState}
    \end{subfigure}

    \caption{Illustration of agent's trajectory under different model defects and baseline plan in the ACAS Xu domain. (a) The agent using the base model completed the task successfully without any collisions. (b-d) Collisions detected at $t=2500$ when the agent operates using models with inaccuracies, despite the existence of a safe plan to complete the task.}
    \label{fig:instrVSmodels}
\end{figure*}

\begin{table*}[t]
    \centering
    \resizebox{\textwidth}{!}{
    \begin{tabular}{|l|l|c|c|c|c|c|}
        \hline
        & & 
        \multicolumn{4}{c|}{\textbf{Execution Anomalies Detected}} & \\
        \cline{3-6}
        \textbf{Domain} & \textbf{Technique} & \textbf{Base} & \textbf{Inacc. State} & \textbf{Inacc. Reward} & \textbf{Both} &  \textbf{State Coverage} \\
        \hline
       \multirow{2}{*}{ACAS Xu} 
& LLM & $\mathbf{4890.2\pm237.6}$ & $\mathbf{5545.0\pm0.0}$ & $\mathbf{5578.0\pm0.0}$ & $\mathbf{5545.0\pm0.0}$ &  $1.0e^{-4} \pm 0.0$ \\
& \textbf{\toolname (10 seeds) } & $483.0 \pm 53.3$& $139.9\pm9.4$ & $138.7\pm12.2$ & $115.5\pm12.05$& $\mathbf{8.1e^{-3}\pm 4.8e^{-4}}$\\
& \textbf{\toolname (20 seeds)}  & $493.4 \pm 45.9$ & $142.5\pm9.59$ & $137.4\pm13.6$ & $115.45\pm10.63$ & $\mathbf{8.2e^{-3}\pm 6.8e^{-4}}$\\
\hline
\multirow{2}{*}{Coop Navi} 
& LLM & $137.0\pm0.0$ & $4603.0\pm0.0$ & $3882.0\pm0.0$ & $9150.0\pm0.0$& $9.9e^{-9} \pm 0.0$\\
& \textbf{\toolname (10 seeds) } & $\mathbf{139.0\pm12.0}$ & $\mathbf{4612.6\pm52.3}$ & $3879.7\pm46.5$  & $\mathbf{9177.1\pm36.4}$ & $\mathbf{9.9e^{-9}\pm 1.1e^{-10}}$\\
& \textbf{\toolname (20 seeds)}  & $\mathbf{138.6\pm10.9}$ & $\mathbf{4615.6\pm1439.6}$ & $\mathbf{3893.2\pm1229.2}$ & $\mathbf{9203.6\pm2864.02}$ & $\mathbf{9.9e^{-9}\pm 1.4e^{-10}}$\\
\hline
\multirow{2}{*}{Flappy Bird} 
& LLM & $\mathbf{7885.0\pm0.0}$ & $9350.0\pm0.0$ & $9578.0\pm0.0$ & $8294.0\pm0.0$ & $62.0 \pm 0.0$\\
& \textbf{\toolname (10 seeds)}  & $1237.6\pm156.8$ & $\mathbf{9355.2\pm80.05}$ & $\mathbf{9605.0\pm97.4}$ & $\mathbf{8307.8\pm101.17}$ & $\mathbf{99.9 \pm 0.3}$ \\
& \textbf{\toolname (20 seeds)}  & $1194.6\pm166.8$ & $9334.85\pm93.4$ & $\mathbf{9582.1\pm110.1}$ & $\mathbf{8285.65\pm80.8}$ & $\mathbf{99.9 \pm 0.2}$ \\
\hline
\multirow{2}{*}{Lava} 
& LLM & $\mathbf{8793.0\pm0.0}$ & $\mathbf{9467.0\pm0.0}$ & $\mathbf{9467.0\pm0.0}$ & $\mathbf{9467.0\pm0.0}$ & $ \mathbf{9.5e^{-5} \pm 0.0} $\\
& \textbf{\toolname (10 seeds)}  & $6815.0 \pm 334.1$ & $6815.0 \pm 334.1$ & $6815.0 \pm 334.1$ & $6815.0 \pm 334.1$ & $8.9e^{-5} \pm 2.4e^{-5} $ \\
& \textbf{\toolname (20 seeds)}  & $6704.8 \pm 303.9$ & $6726.5 \pm 317.1$ & $6726.5 \pm 317.1$ & $6726.5 \pm 317.1$ & $8.9e^{-5} \pm2.1e^{-5}$\\
\hline
\multirow{2}{*}{BipedalWalker} 
& LLM & $\mathbf{8101.2\pm 0}$  & $10000\pm 0$ & $10000\pm 0$ & $10000\pm 0$ & $2.4e^{-3} \pm 0.0 $  \\
& \textbf{\toolname (10 seeds)}  & $7880\pm211.4$ & $\mathbf{10000\pm 0}$ & $\mathbf{10000\pm 0}$ & $\mathbf{10000\pm 0}$ &  $\mathbf{3.0e^{-3} \pm 1.0e^{-4}}$ \\
& \textbf{\toolname (20 seeds)}  & $7890\pm166.8$ & $\mathbf{10000\pm 0}$ & $\mathbf{10000\pm 0}$ & $\mathbf{10000\pm 0}$ &  $\mathbf{3.0e^{-3} \pm 1.0e^{-4}}$  \\
\hline
\end{tabular}
}
\caption{Average number of execution anomalies detected with \toolname-generated and LLM-generated environment-task configurations, along with their state coverages. ``Both'' refers to an agent model of the domain with inaccurate state representation and inaccurate reward function. Best values in each domain are indicated in \textbf{bold}.}
\label{tab:LLM}
\vspace{-2ex}
\end{table*}

\section{Results and Discussion}
\label{sec:results}

\paragraph{Discovering execution anomalies} To answer RQ1, we compare \toolname with baselines, based on the state coverage and the number of execution anomalies that can be detected on the \emph{base models}. We focus this evaluation only on the base models since the existing works only consider them. \toolname generates $10,000$ environment-task configurations, per seed, for evaluation in each domain. The results of MDPFuzz and CureFuzz are averaged over five seeds, as described in their papers. The baselines start with a set of seed configurations and mutate them with 12~hrs timeout per seed, generating significantly more than $10,000$~configurations for evaluation. 

Since multiple environment-task configurations may result in the same execution anomaly, we report both unique and total anomalies detected by each technique in Table~\ref{tab:crashes}, along with the state coverage. \toolname outperforms the baselines in terms of unique execution anomalies, often by a wide margin, and achieves a higher state coverage across majority of the domains. Unlike the baselines that have a large gap between total and unique anomalies in some domains, \toolname consistently yields nearly identical values for both, indicating more precise and less redundant detection. The baselines tend to detect higher total anomalies as they may repeatedly trigger the same anomaly across many configurations until timeout. In contrast, our approach uses a fixed number of configurations and identifies a greater number of unique anomalies. Increasing the number of seeds for \toolname seems to have a minimal effect on performance, indicating that the technique is stable and effective even with fewer runs.

\paragraph{Agent Errors and Environment Errors} To answer RQ2, we apply \toolname to test agents with different types of model errors. In each domain, using the $10,000$ environment-task configurations generated by \toolname per seed, we first determine how many of these are infeasible scenarios (environment errors) using \toolname's heuristic search with a timeout of 1.5~hrs per configuration. For each domain, in each of the feasible scenarios, we test the agents with different model fidelities. If the agent is unable to complete a feasible task, then it is treated as an agent error. 

Table~\ref{tab:agent_env_errors} reports the average number of agent and environment errors detected, along with their standard deviation. The results show that \toolname consistently generates mostly feasible environment-task configurations across domains, especially in ACAS Xu and Coop Navi. This can sometimes happen since we do not explicitly set the number of feasible and infeasible tasks that must be generated. We only aim to generate diverse environment-task scenarios with Latin Hypercube sampling. The number of agent errors is significantly higher in models with injected defects, indicating that \toolname effectively exposes model-specific errors. Interestingly, in ACAS Xu, there are \emph{fewer} agent errors associated with inaccurate models. Our analysis of the performance and error logs revealed that operating under erroneous models allowed the agent to fly \emph{faster} and cover more distance, thereby avoiding collisions.  Due to the large state space in Flappy Bird, our search timed out for $\sim\!7700$~scenarios. 
Nevertheless, \toolname successfully detects many environment errors and agent errors in this domain.

We do not report the results on Bipedal domain since our heuristic search does not support non-determinism. While we can calculate the execution anomalies since it only requires generating environment-task configurations for agent evaluation, i.e. observing whether the agent succeeded or failed, we cannot distinguish between environment errors and agent errors, since our search does not support finding a plan for settings that are \emph{not} fully deterministic.

Figure~\ref{fig:instrVSmodels} shows a visualization of the agent's trajectory, following its policy, under different model fidelities, along with the plan found by our baseline planner in the ACAS Xu domain. With the base model, the agent successfully avoids collision. However, the agent is unable to avoid collisions when operating under erroneous models, even though a collision-free plan exists, as identified by our baseline planner. The figure highlights that even minor model inaccuracies can result in collisions, emphasizing the importance of accurate modeling and exhaustive testing in safety-critical settings. These results highlight the \toolname's ability to stress-test models and detect execution anomalies.

\paragraph{Generating configurations using LLMs} To answer RQ3, we investigate the efficiency of Large Language Models (LLMs) in generating environment-task configurations for agent testing. Specifically, we prompted the LLM to generate environment-task configurations where the agent will likely fail, since they define the boundaries of agent operation. Our prompt to GPT-4o included a description of the agent and its capabilities, and it was tasked with generating $10,000$ environment-task configurations. We then assessed the task feasibility using the same Oracle baseline planner described earlier. Table~\ref{tab:LLM} compares the results of agent evaluation in \toolname-generated environment-task configurations with those generated by GPT-4o. 

The LLM-generated configurations consistently uncover a higher number of anomalies but result in low state coverage, suggesting narrow or adversarial sampling. This result also indicates that agent errors may be sparsely distributed in the state space. This insight is valuable because it suggests that the system may be robust in general, but vulnerable in specific contexts. The results show that designing test cases that are tailored to the agent capabilities may be useful, when the information is available. While LLM-based generation is successful in exposing many anomalies in these domains, \toolname strikes a balance between high anomaly detection and diverse state coverage, offering a more reliable evaluation framework for model robustness. 

\begin{table}[t]
    \renewcommand{\arraystretch}{1.1}
    \centering
    \resizebox{\columnwidth}{!}{
    \begin{tabular}{|l|l|c|c|c|}
        \hline
        \textbf{Domain} & \textbf{Baseline Planner} & \textbf{\#Feasible} & \textbf{\#Infeasible} & \textbf{\#Timeout} \\
        \hline
\multirow{2}{*}{ACAS Xu} 
& BFS    & $9975$ & $3$ & $22$ \\
& \textbf{\toolname search} & $\mathbf{9977}$ & $\mathbf{22}$ & $\mathbf{1}$ \\
\hline
\multirow{2}{*}{Coop Navi} 
& BFS  & $9872$ & $0$ &$128$  \\
& \textbf{\toolname search}  & $\mathbf{9939}$ & $\mathbf{34}$ & $\mathbf{27}$  \\
\hline
\multirow{2}{*}{Flappy Bird} 
& BFS  & $1411$ & $20$ &$8569$  \\
& \textbf{\toolname search} & $\mathbf{1472}$ & $\mathbf{817}$ & $\mathbf{7711}$ \\
\hline
\multirow{2}{*}{Lava} 
& BFS   & $3062$ & $6938$ & $0$ \\
& \textbf{\toolname search}  & $\mathbf{3062}$ & $\mathbf{6938}$ & $\mathbf{0}$ \\
        \hline
    \end{tabular}
    }
    \caption{Number of feasible and infeasible tasks identified by Breadth First Search (BFS) and our proposed heuristic search. Timeout indicates the \#environment-task configurations where the search terminated due to 30 min time-limit.}
    \label{tab:search}
\vspace{-2ex}
\end{table}

\paragraph{Efficiency of BFS as a baseline planner} To answer RQ4,  we compare the \toolname's search with that of Breadth First search (BFS) as a baseline planner. We evaluate their performance based on the number of environment-task settings they could find a plan (feasible), number of environment-task settings where they could determine that no valid plan exists (infeasible), and the number of environment-task settings where they terminated due to a 30-minute per setting cutoff. Table~\ref{tab:search} summarizes these results. Across all domains, the proposed \toolname search performs on-par or better than BFS, with fewer timeouts. We do not report the results on Bipedal domain since both the search techniques cannot solve for domains that are not fully deterministic. While any search technique can be used as a baseline planner, the results show that the ability of the planner to quickly solve large settings is critical for a faithful attribution of environment and agent errors.

\paragraph{Limitations and Results Validity}
\toolname inherits the limitations of the search-based planner, including state space explosion in high-dimensional, continuous environments. While \toolname addresses this by using the binning strategy, it may not always be able to identify a satisficing solution in reasonable time, as observed for the Flappy Bird domain. Similarly, \toolname's heuristic search assumes that Oracle planner deterministically updates the environment state after applying a generated set of actions. In domains such as BipedalWalker, which are implemented using physics engines (e.g., Box2D) the accumulation of floating-point errors causes non-determinism, limiting \toolname's ability to find a satisficing solution in reasonable time. Overcoming these challenges is a promising direction for future research.   

Inspired by the ``threats to validity'' discussions in software testing, we outline key factors that affect the validity of our results and how we overcome them. 
We address the threat to internal validity (threats related to factors within
the experimental design) by reusing the publicly available implementations of the domains and baselines.  Additionally, we run all experiments with multiple random seeds and report averaged results to account for stochastic variation. 
We mitigate the threat to external validity (threats related to the generalizability of our results) by considering a diverse set of continuous and discrete domains. We also evaluate \toolname across different types of models and compare it's performance with two state-of-the-art baselines and LLM. 

\section{Related Work}
\paragraph{Automated testing of autonomous systems}
It is practically infeasible to manually evaluate an autonomous system on all possible scenarios it may encounter in the deployed environment, which motivated automating the testing process~\cite{karimoddini2022automatic}. Fuzz testing is a software testing technique that involves testing the system on a large number of random inputs to uncover bugs, crashes, security vulnerabilities, or other unexpected behavior. Prior works that apply fuzz testing to autonomous agents, such as CureFuzz~\cite{he2024curiosity} and MDPFuzz~\cite{pang2022mdpfuzz}, focus on generating diverse input scenarios, using techniques such as curiosity-driven exploration to uncover edge cases that lead to agent failure.  
Another line of work uses search-based methods to generate adversarial test cases in the form of environment-task configurations where the agent will likely fail, which can be integrated with fuzz-testing~\cite{tappler2022search,tappler2024learning}. The search-based methods require access to the agent's internal model to generate test cases and therefore cannot be applied to black-box systems. 
Alternatively, differential testing has also been used to evaluate model updates to the agent~\cite{nayyar2022differential}. All these approaches primarily focus on detecting model-specific failures without explicitly addressing the feasibility of tasks within the environment itself. Further, many of them require access to the agent's internal model. In contrast, \toolname can detect both environment errors and agent errors in black-box systems.

\paragraph{Model cards}
To improve the transparency of machine learning systems,  model cards have been introduced to document the training and evaluation settings~\cite{mitchell2019model,crisan2022interactive}.  Our testing framework enables principled, exhaustive testing of autonomous agents, providing the data to create model cards for autonomous agents. 

\section{Summary and Future Work}
We present \toolname, a novel framework to evaluate both agent reliability and environmental suitability for deployment. Our evaluation shows that \toolname outperforms the state-of-the-art by detecting significantly more number of unique execution anomalies, attributing anomalies to agent or environment errors, and uniformly covering the state space of the environments. 
A promising direction for future research is to address the limitations of our search-based oracle planner, and extend it to support testing in stochastic environments. This will enable the application of the tool to real-world complex domains such as autonomous vehicles, robotics, and healthcare where deployment decisions must balance model performance with environmental constraints.


\section{Acknowledgment}
This work was supported in part by NSF award 2416459.

\bibliography{relatedwork}
\end{document}